\documentclass[a4paper]{article}

\usepackage{INTERSPEECH2019}

\title{Nonparallel Emotional Speech Conversion}
\name{Jian Gao $^1$, Deep Chakraborty $^2$, Hamidou Tembine $^1$, Olaitan Olaleye $^3$}
\address{
$^1$ Department of Computer Science and Engineering, New York University, USA\\
$^2$ College of Information and Computer Sciences, University of Massachusetts Amherst, USA\\
$^3$ Signify (formerly Philips Lighting) Research, North America, USA
}
\email{\{jg4631, tembine\}@nyu.edu, dchakraborty@cs.umass.edu, olaitan.olaleye@signify.com}

\begin{document}

\maketitle
\begin{abstract}
We propose a nonparallel data-driven emotional speech conversion method. It enables the transfer of emotion-related characteristics of a speech signal while preserving the speaker's identity and linguistic content. Most existing approaches require parallel data and time alignment, which is not available in many real applications. We achieve nonparallel training based on an unsupervised style transfer technique, which learns a translation model between two distributions instead of a deterministic one-to-one mapping between paired examples. The conversion model consists of an encoder and a decoder for each emotion domain. We assume that the speech signal can be decomposed into an emotion-invariant content code and an emotion-related style code in latent space. Emotion conversion is performed by extracting and recombining the content code of the source speech and the style code of the target emotion. We tested our method on a nonparallel corpora with four emotions. The evaluation results show the effectiveness of our approach.
\end{abstract}
\noindent\textbf{Index Terms}: Emotional Speech Conversion, Non-parallel training, Style Transfer, Autoencoder, GANs

\section{Introduction}
\label{sec:intro}
Voice transformation (VT) is a technique to modify some properties of human speech while preserving its linguistic information. VT can be applied to change the speaker identity, i.e., voice conversion (VC)~\cite{mohammadi2017overview}, or to transform the speaking style of a speaker, such as emotion and accent conversion~\cite{zhao2018accent}. In this work, we will focus on emotion voice transformation. The goal is to change emotion-related characteristics of a speech signal while preserving its linguistic content and speaker identity. Emotion conversion techniques can be applied to various tasks, such as enhancing computer generated speech, hiding negative emotions for people, helping film dubbing, and creating more expressive voice messages on social media.

%{\color{blue}Numerous use-cases exist for this solution including customer service representative conversation masking, financial earnings report conference calls emotion masking, conference room solutions or Instagram-like emotion-filters for speech in social media applications. \textbf{shorter?}}.

Traditional VC approaches cannot be applied directly because they change speaker identity by assuming pronunciation and intonation to be a part of the speaker-independent information. Since the speaker's emotion is mainly conveyed by prosodic aspects, some studies have focused on modelling prosodic features such as pitch, tempo, and volume \cite{wang2012emotional,wang2014multi}. In \cite{xue2018voice}, a rule-based emotional voice conversion system was proposed. It modifies prosody-related acoustic features of neutral speech to generate different types of emotions. A speech analysis-synthesis tool STRAIGHT \cite{kawahara1999restructuring} was used to extract fundamental frequency ($F_0$) and power envelope from raw audio. These features were parameterized and modified based on Fujisaki model \cite{fujisaki1984analysis} and target prediction model \cite{xue2016study}. The converted features were then fed back into STRAIGHT to re-synthesize speech waveforms with desired emotions. However, this method requires temporal aligned parallel data that is difficult to obtain in real applications; and the accurate time alignment needs manual segmentation of the speech signal at phoneme level, which is very time consuming.
%Thus this method has limitations in real-world situations.
%The manual interaction and calibration is also very time consuming.
%On the other side, automatic time alignment algorithm like dynamic time warping (DTW) may cause performance degradation due to mismatching.
% time consuming and tedious task

%it's harder to get parallel emotional speech corpora (dataset XXX) than VC.
%need professional actors, since in normal life, it's rare to find angry sentences pronounced in a happy style
To address these issues, we propose a nonparallel training method. Instead of learning one-to-one mapping between paired emotional utterances $(x_1, x_2)$, we switch to training a conversion model between two emotional domains $(\mathcal{X}_1, \mathcal{X}_2)$.

Inspired by disentangled representation learning in image style transfer \cite{gatys2016image,Huang_2018_ECCV}, we assume that each speech signal $x_i \in \mathcal{X}_i$ can be decomposed into a content code $c \in \mathcal{C}$ that represents emotion-invariant information and a style code $s_i \in \mathcal{S}_i$ that represents emotion-dependent information. $\mathcal{C}$ is shared across domains and contains the information we want to preserve. $\mathcal{S}_i$ is domain-specific and contains the information we want to change. In conversion stage, we extract content code of the source speech and recombine it with style code of the target emotion. A generative adversarial network (GAN) \cite{goodfellow2014generative} is added to improve the quality of converted speech. Our approach is nonparallel, text-independent, and does not rely on any manual operation.
%It can be trained on a small amount of utterances ($\sim$ 8 min per emotion).

%We employ gated convolutional neural networks (CNNs) ~\cite{dauphin2017language} to model the speech representations. This allows the autoencoder to capture the long-range dependencies in speech. The conversion model is trained with a weak cycle consistency loss ~\cite{Zhu_2017_ICCV, huang2018multimodal}, which helps to create pseudo pairs from nonparallel data. And and keep it indistinguishable from the real ones.

We evaluated our approach on IEMOCAP \cite{busso2008iemocap} for four emotions: angry, happy, neutral, sad; which is widely studied in emotional speech recognition literature \cite{8682541}. To our knowledge, this is the first attempt of nonparallel emotion conversion on this dataset, though synthetic feature representations of emotional speech were proposed in \cite{DBLP:conf/interspeech/SahuGE18}. We evaluate the model's conversion ability by the percentage change from source emotion to target emotion.
%It is a nonparallel dataset widely used in emotional speech recognition and analysis.
A subjective evaluation on Amazon MTurk with hundreds of listeners was conducted. It shows our model can effectively change emotions and retain the speaker identity.

The rest of the paper is organized as follows: Section \ref{sec:related} presents the relation to prior work. Section \ref{sec:method} gives a detailed description of our model. Experiments and evaluation results are reported in Section \ref{sec:exp}. Finally, we conclude in Section \ref{sec:con}.

\section{Related Work}
\label{sec:related}

\subsection{Emotion-related features}
Previous emotion conversion methods directly modify parameterized prosody-related features that convey emotions. The use of Gaussian mixture models (GMM) for spectrum transformation was first proposed in \cite{kawanami2003gmm}.
%[12] introduced a data-driven emotion conversion system that combines independent parameter transformation techniques including HMM-based $F_0$ generation, $F_0$ segment selection, duration conversion and GMM-based spectral conversion. However, it requires large amounts of parallel data.
A recent work ~\cite{xue2018voice} explored four types of acoustic features: $F_0$ contour, spectral sequence, duration and power envelope, and investigated their impact on emotional speech synthesis
% by controlled feature replacement.
The authors found that $F_0$ and spectral sequence are the dominant factors in emotion conversion, while power envelope and duration alone has little influence. They further claimed that all emotions can be synthesized by modifying the spectral sequence, but did not provide a method to do it. In this paper, we focus on learning the conversion models for $F_0$ and spectral sequence.
%Since changing duration and power envelope requires manually segmenting the phoneme boundaries of vowels and consonants, we leave it for future work.

\subsection{Nonparallel training approaches}
%In parallel setting, the training data is limited to a small amount of predefined sentences, which may impair the generalization ability.
%It is a time consuming and painstaking task to collect parallel data. Training on well-aligned parallel data is easy, but collecting such data is hard.
Parallel data means utterances with the same linguistic content but varying in aspects to be studied. Since parallel data is hard to collect, nonparallel approaches have been developed. Some borrow ideas from image-to-image translation \cite{NIPS2017_6672} and create GAN models \cite{goodfellow2014generative} suitable for speech, such as VC-VAW-GAN \cite{Hsu2017}, SVC-GAN \cite{Kaneko2017}, VC-CycleGAN \cite{fang2018high,DBLP:journals/corr/abs-1904-04631}, VC-StarGAN \cite{DBLP:conf/interspeech/SahuGE18}. Another trend is based on auto-regressive models like WaveNet \cite{van2016wavenet}. Although it can train directly on raw audio without feature extraction, the heavy computational load and huge amount of training data required is not affordable for most users.

%Our approach only needs a small amount of training data ($\sim$ 8 min per emotion), and does not rely on any transcripts, manual operation, or preprocessing step.
%Our method differs in that; external data

\subsection{Disentangled representation learning}
Our work draws inspiration from recent studies in image style transfer. A basic idea is to find disentangled representations that can independently model image content and style. It is claimed in~\cite{gatys2016image} that a Convolutional Neural Network (CNN) is an ideal representation to factorize semantic content and artistic style. They introduced a method to separate and recombine content and style of natural images by matching feature correlations in different convolutional layers. For us, the task is to find disentangled representations for speech signal that can split emotion from speaker identity and linguistic content.

\section{Method}
\label{sec:method}

\subsection{Assumptions}
%Emotions are nonlinguistic information not generally controlled by speakers.
The research on human emotion expression and perception has two major conclusions.
First, human emotion perception is a multi-layered process. It's figured out that humans do not perceive emotion directly from acoustic features, but through an intermediate layer of semantic primitives \cite{huang2008three}. They introduced a three-layered model and learnt the connections by a fuzzy inference system. Some researchers found that adding middle layers can improve emotion recognition accuracy \cite{Li2016MultilingualSE}. Based on this finding, we suggest the use of multilayer perceptrons (MLP) to extract emotion-related information in speech signals.

Second, the emotion generation process of human speech follows the opposite direction of emotion perception. This means the encoding process of the speaker is the inverse operation of the decoding process of the listener. We assume that emotional speech generation and perception share the same representation methodology, that is, the encoder and decoder are inverse operations with mirror structures.

Let $x_1 \in \mathcal{X}_1$ and $x_2 \in \mathcal{X}_2$ be utterances drawn from two different emotional categories. Our goal is to learn a mapping between two distributions $p(x_1)$ and $p(x_2)$. Since the joint distribution $p(x_1, x_2)$ is unknown for nonparallel data, the conversion models $p(x_1|x_2)$ and $p(x_2|x_1)$ cannot be directly estimated. To solve this problem, we make two assumptions: \\
(i). The speech signal can be decomposed into an emotion-invariant content code and an emotion-dependent style code; \\
(ii). The encoder $E$ and decoder $G$ are inverse functions.

\subsection{Model}
\begin{figure}[htb]
\center
\includegraphics[width=0.4\textwidth]{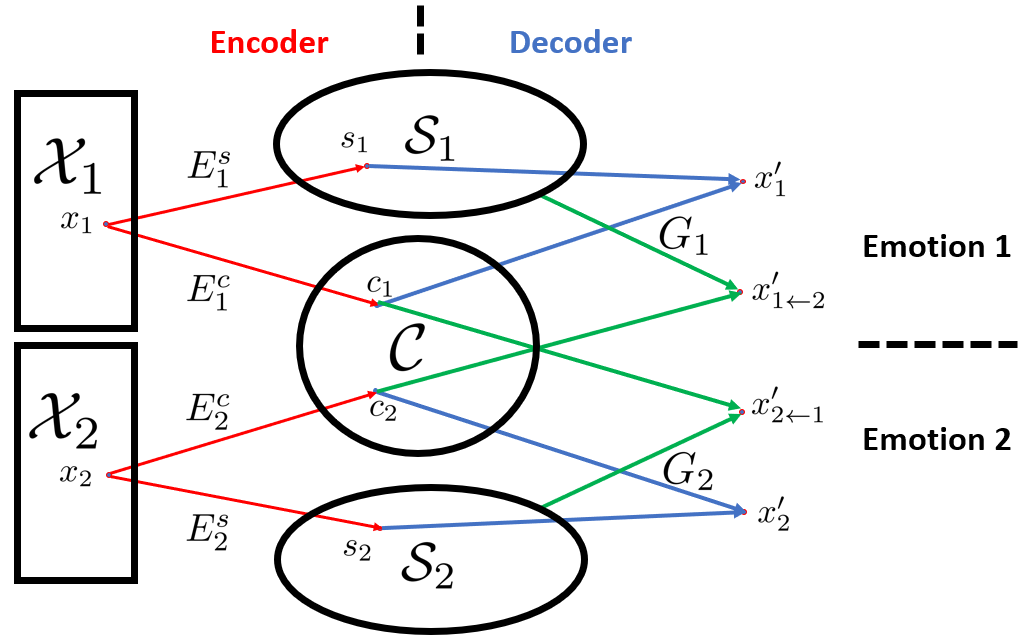}
\caption{Speech autoencoder model with partially shared latent space. Speech with emotion $i$ is decomposed into an emotion-specific space $\mathcal{S}_i$ and a shared content space $\mathcal{C}$. Corresponding speech $(x_1,x_2)$ are encoded to the same content code.
%Emotion conversion is obtained by recombining the content code of the input utterance and the style code randomly sampled from the target emotion space.
}
\label{autoencoder}
\end{figure}

Fig. \ref{autoencoder} shows the generative model of speech with a partially shared latent space. A pair of corresponding speech $(x_1, x_2)$ is assumed to have a shared latent code $c \in \mathcal{C}$ and emotion-related style codes $s_1 \in \mathcal{S}_1, s_2 \in \mathcal{S}_2$. For any emotional speech $x_i$, we have a deterministic decoder $x_i = G_i(c_i,s_i)$ and its inverse encoders $c_i = E_i^c(x_i)$, $s_i = E_i^s(x_i)$. To convert emotion, we just extract and recombine the content code of the source speech with the style code of the target emotion.
\begin{equation}
\begin{aligned}
x_{1\leftarrow2}' = G_1(c_2, s_1) = G_1(E_2^c(x_2), s_1) \\
x_{2\leftarrow1}' = G_2(c_1, s_2) = G_2(E_1^c(x_1), s_2)
\end{aligned}
\end{equation}
It should be noted that the style code $s_i$ is not inferred from one utterance, but learnt from the entire emotion domain. This is because the emotion style from a single utterance is ambiguous and may not capture the general characteristics of the target emotion. It makes our assumption slightly different from the cycle consistent constraint ~\cite{Zhu_2017_ICCV}, which assumes that an example converted to another domain and converted back should remain the same as the original, i.e., $x_{1\leftarrow2\leftarrow1}'' = x_1$. Instead, we apply a semi-cycle consistency in the latent space by assuming that $E_1^c(x_{1\leftarrow2}') = c_1$ and $E_1^s(x_{1\leftarrow2}') = s_1$.

\begin{figure}[htb]
\center
\includegraphics[width=0.4\textwidth]{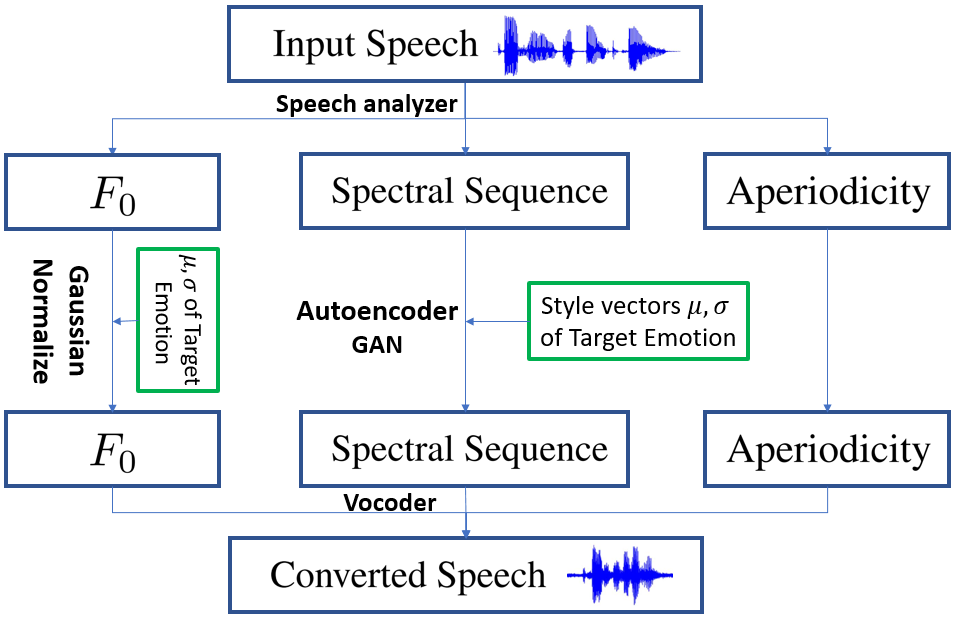}
\caption{Overview of nonparallel emotion conversion system}
\label{model}
\end{figure}

% Traditional emotional speech analysis mainly focuses on four types of acoustic features: fundamental frequency ($F_0$), spectral sequence, time duration and energy envelope. It was found in ~\cite{xue2018voice} that only $F_0$ and spectral sequence have significant influence, while the other two require manual segmentation and have little impact on changing emotions. Therefore we focus on learning the conversion model for $F_0$ and spectral sequence.
Fig. \ref{model} shows an overview of our nonparallel emotional speech conversion system. The features are extracted and recombined by WORLD \cite{morise2016world} and converted separately. We modify $F_0$ by linear transform to match statistics of the fundamental frequencies in the target emotion domain. The conversion is performed by log Gaussian normalization
\begin{equation}
f_2 = \exp((\log f_1 - \mu_1)\cdot\frac{\sigma_2}{\sigma_1} + \mu_2)
\label{eq:f0}
\end{equation}
where $\mu_i, \sigma_i$ are the mean and variance obtained from the source and target emotion set. Aperiodicity (AP) is mapped directly since it does not contain emotion-related information.

For spectral sequence, we use low-dimensional representation in mel-cepstrum domain to reduce complexity. Kameoka \cite{DBLP:conf/slt/KameokaKTH18} shows 50 MCEP coefficients are enough to synthesize full-band speech without quality degeneration. Spectra conversion is learnt by the autoencoder model in Fig. \ref{autoencoder}. The encoders and decoders are implemented with gated CNN \cite{dauphin2017language}. In addition, a GAN module is added and trained by robust optimization \cite{Gao18} to produce realistic spectral frames. Our model has 4 subnetworks $E^c, E^s, G, D$, in which $D$ is the discriminator in GAN to distinguish real samples from machine-generated samples.

\subsection{Loss functions}
\begin{figure}[htb]
\includegraphics[width=0.47\textwidth]{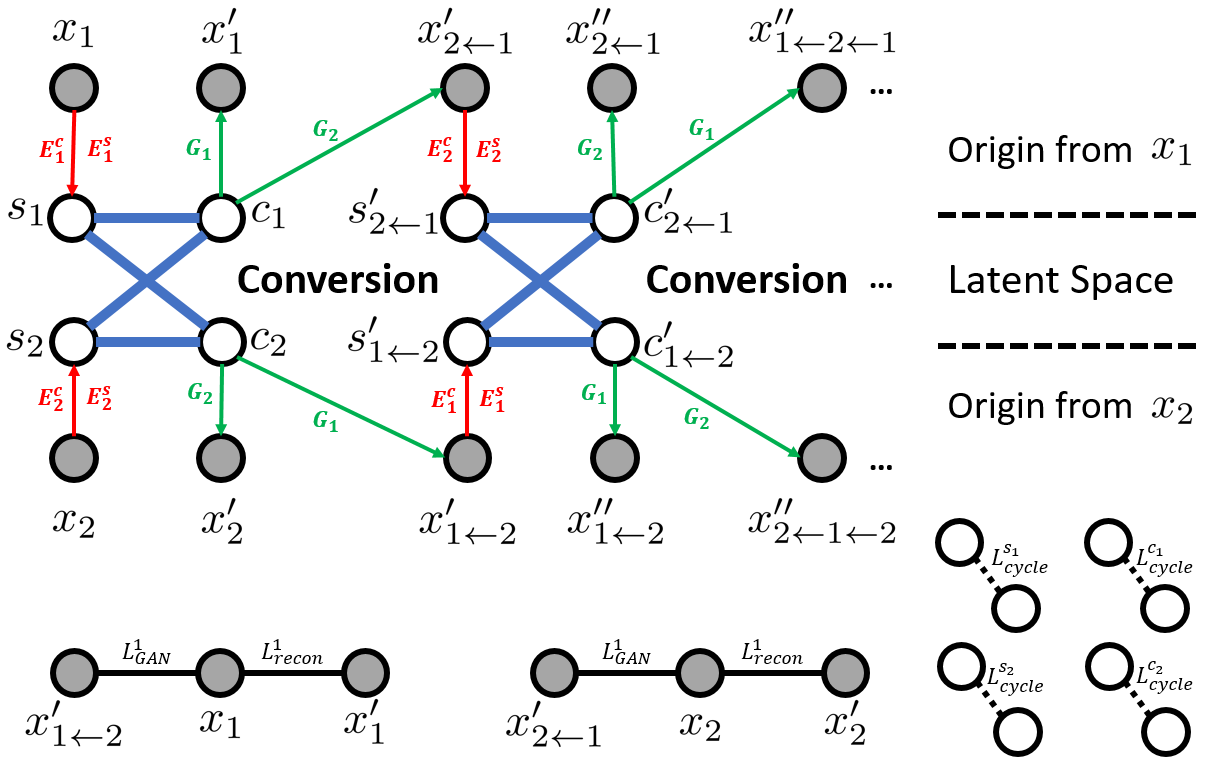}
\caption{Train on multiple loss functions}
\label{loss}
\end{figure}

We jointly train the encoders, decoders and GAN's discriminators with multiple losses displayed in Fig. \ref{loss}. To keep encoder and decoder as inverse operations, we apply reconstruction loss in the direction $x_i \rightarrow (c_i, s_i) \rightarrow x_i'$. The spectral sequence should not change after encoding and decoding.
\begin{equation}
L_{recon}^{x_i} = \mathbb{E}_{x_i}(\| x_i - x_i' \|_1), \quad x_i' = G_i(E_i^c(x_i), E_i^s(x_i))
\end{equation}

In our model, the latent space is partially shared. Thus the cycle consistency constraint ~\cite{Zhu_2017_ICCV} is not preserved, i.e., $x_{1\leftarrow2\leftarrow1}'' \neq x_1$. We apply a semi-cycle loss in the coding direction $c_1 \rightarrow x_{2\leftarrow1}' \rightarrow c_{2\leftarrow1}'$ and $s_2 \rightarrow x_{2\leftarrow1}' \rightarrow s_{2\leftarrow1}'$.
\begin{equation}
\begin{aligned}
L_{cycle}^{c_1} = \mathbb{E}_{c_1, s_2} (\| c_1 - c_{2\leftarrow1}' \|_1), \quad c_{2\leftarrow1}'=E_2^c(x_{2\leftarrow1}') \\
L_{cycle}^{s_2} = \mathbb{E}_{c_1, s_2} (\| s_2 - s_{2\leftarrow1}' \|_1), \quad s_{2\leftarrow1}'=E_2^s(x_{2\leftarrow1}')
\end{aligned}
\end{equation}
Moreover, we add a GAN module to improve the speech quality. The converted samples should be indistinguishable from the real samples in the target emotion domain. GAN loss is computed between $x_{i\leftarrow j}'$ and $x_i$, $(i \neq j)$.
\begin{equation}
L_{GAN}^i = \mathbb{E}_{c_j, s_i}[\log(1-D_i(x_{i\leftarrow j}'))] + \mathbb{E}_{x_i}[\log D_i(x_i)]
\end{equation}
The full loss is the weighted sum of $L_{recon}$, $L_{cycle}$, $L_{GAN}$.
\begin{equation}
\begin{aligned}
\min_{E_1^c,E_1^s,E_2^c,E_2^s, G_1,G_2}\max_{D_1,D_2} L(E_1^c, E_1^s, E_2^c, E_2^s, G_1, G_2, D_1, D_2) \\
= \lambda_s (L_{cycle}^{s_1} + L_{cycle}^{s_2}) + \lambda_c (L_{cycle}^{c_1} + L_{cycle}^{c_2}) \ \qquad \qquad \\
+ \lambda_x (L_{recon}^1 + L_{recon}^2) + \lambda_g (L_{GAN}^1 + L_{GAN}^2) \qquad \quad
\end{aligned}
\end{equation}
where $\lambda_s, \lambda_c, \lambda_x, \lambda_g$ control the weights of the components.

\begin{figure*}[t!]
\includegraphics[width=1.0\textwidth]{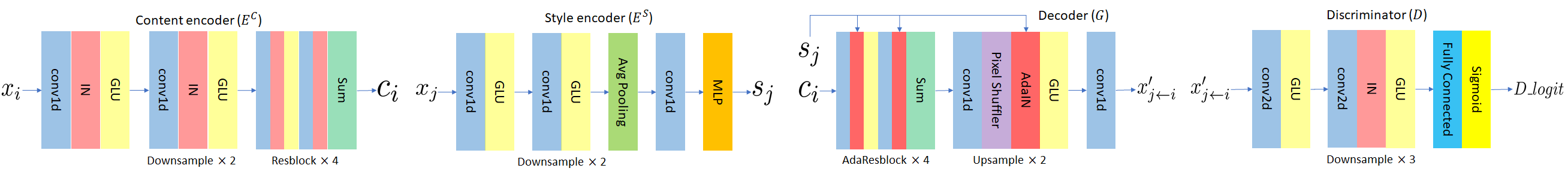}
\caption{The network structure of content encoder, style encoder, decoder, and GAN discriminator.}
\label{fig:NN}
\end{figure*}

\section{Experiments}
\label{sec:exp}
Training emotional speech conversion models often suffers from lack of data. Parallel datasets such as Emo-DB \cite{BurkhardtPRSW05} and RAVDESS \cite{10.1371/journal.pone.0196391} have limited sentence diversity and are difficult to build. Our end-to-end model is trained on raw audio signals of natural speech, and does not rely on paired data or any manual operations. Training set can be collected from daily conversations in everyday life.

\subsection{Experiment setup}
We evaluated the proposed approach on the Interactive Emotional Dyadic Motion Capture database (IEMOCAP)  \cite{busso2008iemocap}. It is organized in five sessions and contains 12 hours of audiovisual data. Each session records natural dialogues between a pair of speakers in scripted and improvised scenarios, in which the emotions are naturally elicited. In this paper, we only consider four emotional categories: 1) angry, 2) happy, 3) neutral, 4) sad. Since the model is not designed to change the speaker identity, experiments are conducted for each speaker independently. We only use the utterances with a clear majority vote regarding the ground truth labels. There are 2754 utterances shared amongst four emotional labels: ang (747), hap (675), neu (788), sad (544). Training and testing sets are non-overlapping utterances randomly selected from the same speaker (80\% for training, 20\% for test). For example in session 1, there are 420 training samples and 108 testing samples for the female speaker. %The emotions in scripted dialogues have strong correlation with the lingual content.

Training samples with fixed length of 128 frames are randomly selected from raw audio sequences. Energy-based voice-activity detection (VAD) is used to remove silent frames. We use WORLD \cite{morise2016world} vocoder to extract fundamental frequencies, spectral sequences (sps) and aperiodicities (aps) from raw audio waveforms sampled at 16KHz. The frame length is 5ms. After coding, we take the first 24 Mel-cepstral coefficients (MCEPs) as feature vectors. Mean and variance of the entire training set are calculated for feature normalization. Testing samples can have arbitrary temporal length, and be converted in real-time.
%of temporal size 128, to feed the autoencoder model.
% deal with variable length of input

\subsection{Network architecture}
The network architecture is illustrated in Fig. \ref{fig:NN}, with details listed in Table \ref{tab:NN}. The autoencoders take 24-dimentional MCEPs as input and learn disentangled representations of content and style. In the content encoder, instance normalization (IN) \cite{DBLP:journals/corr/UlyanovVL16} removes the original feature mean and variance that represent emotional style information. In the style encoder, the emotional characteristics are encoded by a $3$-layer MLP that outputs channel-wise mean and variance $\mu(s), \sigma(s)$. Then they are fed into the decoder to reconstruct MCEP features. The desired emotion is added through an adaptive instance normalization (AdaIN) \cite{Huang_2017_ICCV} layer before activation. This mechanism is similar to the conversion model of $F_0$ in eq. (\ref{eq:f0}).
\begin{equation}
\text{AdaIN}(c,s) = \sigma(s)\Big(\frac{c-\mu(c)}{\sigma(c)}\Big) + \mu(s)
\end{equation}
The encoders and decoders are implemented with 1D-CNNs to capture the temporal dependencies, while the GAN discriminators are implemented with 2D-CNNs to capture the spectra-temporal patterns. Higher resolution data is generated by the pixel shuffler layer in upsample blocks. All networks use gated linear units (GLU) \cite{dauphin2017language} to keep track of sequential information.

\begin{table}[th]
\caption{Network Architecture. C-F-K-S-X indicates convolution layer with filters F, kernel size K, strides S, and shuffle X. IN is instance normalization; all modules use GLU activation. }
\label{tab:NN}
\centering
\begin{tabular}{c|c}
\hline
\multicolumn{2}{c}{\bf{Content Encoder}} \\ \hline\hline
Conv1d, IN, GLU & C-128-15-1 \\ \hline
Downsample1d $\times$ 2 & C-256-5-2, C-512-5-2 \\ \hline
Resblock1d $\times$ 4 & C-512-3-1, content code \\ \hline\hline
\multicolumn{2}{c}{\bf{Style Encoder}} \\ \hline\hline
Conv1d, GLU & C-128-15-1 \\ \hline
Downsample1d without IN $\times$ 2 & C-256-5-2, C-512-5-2 \\ \hline
Downsample1d without IN $\times$ 2 & C-512-3-2 \\ \hline
Adaptive average pooling \\ \hline
Conv1d & C-16-1-1 \\ \hline
MLP: linear $\times$ 2 & flatten, dense output \\ \hline\hline
\multicolumn{2}{c}{\bf{Decoder}} \\ \hline
Adaptive Resblock1d $\times$ 3 & C-512-3-1 \\ \hline
Upsample1d $\times$ 2 & C-512-5-1-2, C-256-5-1-2\\ \hline
Conv1d & C-24-15-1, MCEPs output \\ \hline\hline
\multicolumn{2}{c}{\bf{Discriminator}} \\ \hline
Conv2d & C-128-(3,3)-(1,2) \\ \hline
Downsample2d & C-256-(3,3)-(2,2) \\ \hline
Downsample2d & C-512-(3,3)-(2,2) \\ \hline
Downsample2d & C-1024-(6,3)-(1,2) \\ \hline
Dense layer & sigmoid output (real/fake) \\ \hline
\end{tabular}
\end{table}

\paragraph*{Training details:}
We use Adam optimizer with $\beta_1=0.5$. The learning rate is initialized as $0.0001$ for $D$ and $0.0002$ for $E^c, E^s, G$; it begins with linear decay applied after $150K$ iterations. Weights are chosen as $\lambda_s = \lambda_c = \lambda_g = 1$, $\lambda_x=10$. $E^c, E^s, G$ are trained 2 iterations for each $D$'s iteration in the first $100K$ iterations; after that they are trained equally.

\subsection{Experiment results}
We evaluate the generated speech on three metrics: voice quality, speaker similarity, and the emotion conversion ability.

\begin{figure}[htb]
\center
\includegraphics[width=0.471\textwidth]{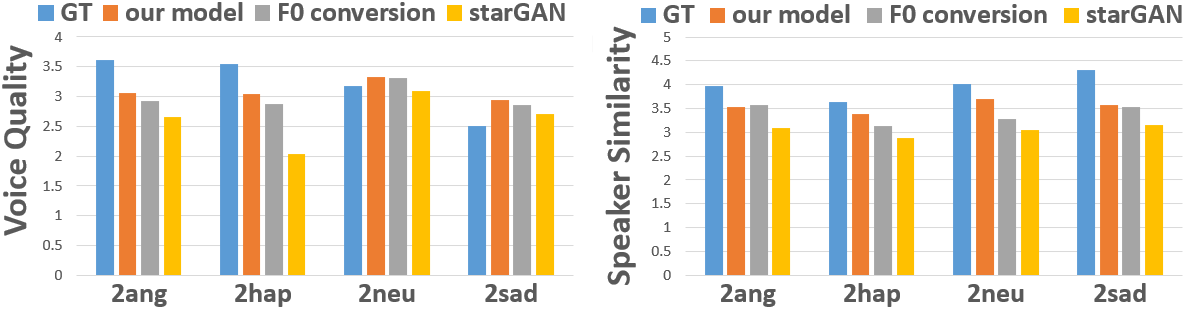}
\caption{MOS for voice quality and speaker similarity. left: voice quality. right: speaker similarity, 2ang means the target emotion is Angry, and compared with originally Angry speech.}
\label{fig:mos}
\end{figure}

\begin{figure}[htb]
\center
\includegraphics[width=0.41\textwidth]{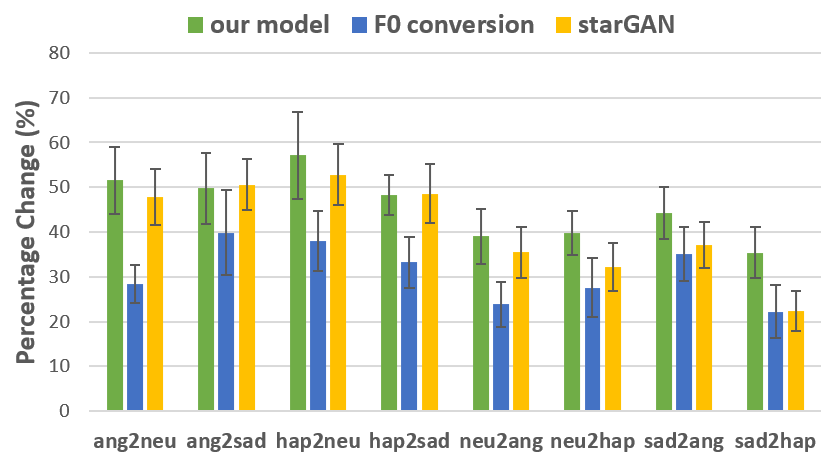}
\caption{Comparison of the emotion conversion ability of our model and the baseline systems: (1) F0 conversion, (2) VC-starGAN \cite{DBLP:conf/slt/KameokaKTH18}. ang2neu is conversion from Angry to Neutral. }
\label{fig:emo}
\end{figure}

\noindent \textbf{Subjective evaluation \ } We perform perception tests on Amazon Mechanical Turk \footnote{https://www.mturk.com}. Each utterance was listened by $5$ random human workers, and each worker can answer at most 5 hits in a single experiment. To evaluate the voice quality and speaker similarity, the listeners were asked to give a 5-scale opinion score (5 for the best, 1 for the worst). The mean opinion score (MOS) is shown in Fig. \ref{fig:mos}. To annotate the emotion state, each listener was asked to choose a label from the source and target emotions. For example in the trial "ang2neu", utterances with label "ang" in IEMOCAP were converted to "neu", and the generated speech was labelled by the majority vote of human annotators. We compute percentage change from the source emotion to the target emotion. Higher value indicates stronger ability of emotional conversion. We choose four emotion pairs with significant differences \cite{BurkhardtPRSW05}. The baseline models are a simple linear $F_0$ conversion system \cite{DBLP:journals/taslp/TaoKL06}, and a neural network model VC-StarGAN \cite{DBLP:conf/slt/KameokaKTH18}. Results are displayed in Fig. \ref{fig:emo}. Details and some converted speech samples are provided at \footnote{https://www.jian-gao.org/emovc}.

\noindent \textbf{Results and Discussion \ } Note that not all utterances can be successfully converted, because some emotions are delivered by linguistic information, an immutable part in our setting. Our model is slightly better than VC-StarGAN in terms of emotion conversion ability (average 48\% vs 44\%) and speaker similarity (average 3.55 vs 3.05). One reason is that VC-StarGAN is designed for voice conversion among different speakers, while our model learns the disentangled representations that can decompose the emotional characteristic and speaker identity. Moreover, VC-StarGAN has poor voice quality in the direction of sad2ang (1.71) and sad2hap (1.81). In \cite{DBLP:conf/slt/KameokaKTH18}, all emotions are trained together, thus it's unfair to the sad domain since it has lower signal to noise ratio, and may amplify the noise when converted to more energetic emotions.

\section{Conclusion}
\label{sec:con}
We proposed a nonparallel emotional speech conversion approach based on style transfer autoencoders. As our model does not require any paired data, transcripts or time alignment, it is easy to apply in real-world situations. To the best of our knowledge, this is the first work on nonparallel emotion conversion using style transfer. Future work includes phonetic duration conversion and designing a general model for unseen speakers.

\section{Acknowledgements}
This research was supported by Signify Research and U.S. Air Force under grant FA9550-17-1-0259.

\clearpage

\bibliographystyle{IEEEtran}
\bibliography{mybib}

% Generated by IEEEtran.bst, version: 1.14 (2015/08/26)
\begin{thebibliography}{10}
\providecommand{\url}[1]{#1}
\csname url@samestyle\endcsname
\providecommand{\newblock}{\relax}
\providecommand{\bibinfo}[2]{#2}
\providecommand{\BIBentrySTDinterwordspacing}{\spaceskip=0pt\relax}
\providecommand{\BIBentryALTinterwordstretchfactor}{4}
\providecommand{\BIBentryALTinterwordspacing}{\spaceskip=\fontdimen2\font plus
\BIBentryALTinterwordstretchfactor\fontdimen3\font minus
  \fontdimen4\font\relax}
\providecommand{\BIBforeignlanguage}[2]{{%
\expandafter\ifx\csname l@#1\endcsname\relax
\typeout{** WARNING: IEEEtran.bst: No hyphenation pattern has been}%
\typeout{** loaded for the language `#1'. Using the pattern for}%
\typeout{** the default language instead.}%
\else
\language=\csname l@#1\endcsname
\fi
#2}}
\providecommand{\BIBdecl}{\relax}
\BIBdecl

\bibitem{mohammadi2017overview}
S.~Mohammadi and A.~Kain, ``An overview of voice conversion systems,''
  \emph{Speech Communication}, vol.~88, pp. 65--82, 2017.

\bibitem{zhao2018accent}
G.~Zhao, S.~Sonsaat, J.~Levis, E.~Chukharev-Hudilainen, and R.~Gutierrez-Osuna,
  ``Accent conversion using phonetic posteriorgrams,'' in \emph{ICASSP}.\hskip
  1em plus 0.5em minus 0.4em\relax IEEE, 2018, pp. 5314--5318.

\bibitem{wang2012emotional}
M.~Wang, M.~Wen, K.~Hirose, and N.~Minematsu, ``Emotional voice conversion for
  mandarin using tone nucleus model--small corpus and high efficiency,'' in
  \emph{Speech Prosody 2012}, 2012.

\bibitem{wang2014multi}
Z.~Wang and Y.~Yu, ``Multi-level prosody and spectrum conversion for emotional
  speech synthesis,'' in \emph{Signal Processing (ICSP)}.\hskip 1em plus 0.5em
  minus 0.4em\relax IEEE, 2014, pp. 588--593.

\bibitem{xue2018voice}
Y.~Xue, Y.~Hamada, and M.~Akagi, ``Voice conversion for emotional speech:
  Rule-based synthesis with degree of emotion controllable in dimensional
  space,'' \emph{Speech Communication}, vol. 102, pp. 54--67, 2018.

\bibitem{kawahara1999restructuring}
H.~Kawahara, I.~Masuda-Katsuse, and A.~De~Cheveigne, ``Restructuring speech
  representations using a pitch-adaptive time--frequency smoothing and an
  instantaneous-frequency-based f0 extraction: Possible role of a repetitive
  structure in sounds1,'' \emph{Speech communication}, vol.~27, no. 3-4, pp.
  187--207, 1999.

\bibitem{fujisaki1984analysis}
H.~Fujisaki and K.~Hirose, ``Analysis of voice fundamental frequency contours
  for declarative sentences of japanese,'' \emph{ASJs Japan (E)}, vol.~5,
  no.~4, pp. 233--242, 1984.

\bibitem{xue2016study}
Y.~Xue and M.~Akagi, ``A study on applying target prediction model to
  parameterize power envelope of emotional speech,'' in \emph{RISP workshop
  NCSP'16}, 2016.

\bibitem{gatys2016image}
L.~Gatys, A.~Ecker, and M.~Bethge, ``Image style transfer using convolutional
  neural networks,'' in \emph{CVPR}.\hskip 1em plus 0.5em minus 0.4em\relax
  IEEE, 2016, pp. 2414--2423.

\bibitem{Huang_2018_ECCV}
X.~Huang, M.~Liu, S.~Belongie, and J.~Kautz, ``Multimodal unsupervised
  image-to-image translation,'' in \emph{The European Conference on Computer
  Vision (ECCV)}, September 2018.

\bibitem{goodfellow2014generative}
I.~Goodfellow, J.~Pouget-Abadie, M.~Mirza, B.~Xu, D.~Warde-Farley, S.~Ozair,
  A.~Courville, and Y.~Bengio, ``Generative adversarial nets,'' in \emph{NIPS},
  2014, pp. 2672--2680.

\bibitem{busso2008iemocap}
C.~Busso, M.~Bulut, C.~Lee, A.~Kazemzadeh, E.~Mower, S.~Kim, J.~Chang, S.~Lee,
  and S.~Narayanan, ``Iemocap: Interactive emotional dyadic motion capture
  database,'' \emph{Language resources and evaluation}, vol.~42, no.~4, p. 335,
  2008.

\bibitem{8682541}
M.~{Neumann} and N.~T. {Vu}, ``Improving speech emotion recognition with
  unsupervised representation learning on unlabeled speech,'' in \emph{ICASSP
  2019 - 2019 IEEE International Conference on Acoustics, Speech and Signal
  Processing (ICASSP)}, May 2019, pp. 7390--7394.

\bibitem{DBLP:conf/interspeech/SahuGE18}
S.~Sahu, R.~Gupta, and C.~Y. Espy{-}Wilson, ``On enhancing speech emotion
  recognition using generative adversarial networks,'' in \emph{Interspeech
  2018, 19th Annual Conference of the International Speech Communication
  Association, Hyderabad, India, 2-6 September 2018.}, 2018, pp. 3693--3697.

\bibitem{kawanami2003gmm}
H.~Kawanami, Y.~Iwami, T.~Toda, H.~Saruwatari, and K.~Shikano, ``Gmm-based
  voice conversion applied to emotional speech synthesis,'' in
  \emph{Eurospeech}, 2003.

\bibitem{NIPS2017_6672}
M.~Liu, T.~Breuel, and J.~Kautz, ``Unsupervised image-to-image translation
  networks,'' in \emph{Advances in Neural Information Processing Systems
  (NIPS)}.\hskip 1em plus 0.5em minus 0.4em\relax Curran Associates, Inc.,
  2017, pp. 700--708.

\bibitem{Hsu2017}
C.-C. Hsu, H.-T. Hwang, Y.-C. Wu, Y.~Tsao, and H.-M. Wang, ``Voice conversion
  from unaligned corpora using variational autoencoding wasserstein generative
  adversarial networks,'' in \emph{Proc. Interspeech 2017}, 2017, pp.
  3364--3368.

\bibitem{Kaneko2017}
T.~Kaneko, H.~Kameoka, K.~Hiramatsu, and K.~Kashino, ``Sequence-to-sequence
  voice conversion with similarity metric learned using generative adversarial
  networks,'' in \emph{Proc. Interspeech 2017}, 2017, pp. 1283--1287.

\bibitem{fang2018high}
F.~Fang, J.~Yamagishi, I.~Echizen, and J.~Lorenzo-Trueba, ``High-quality
  nonparallel voice conversion based on cycle-consistent adversarial network,''
  \emph{arXiv preprint arXiv:1804.00425}, 2018.

\bibitem{DBLP:journals/corr/abs-1904-04631}
\BIBentryALTinterwordspacing
T.~Kaneko, H.~Kameoka, K.~Tanaka, and N.~Hojo, ``Cyclegan-vc2: Improved
  cyclegan-based non-parallel voice conversion,'' \emph{CoRR}, vol.
  abs/1904.04631, 2019. [Online]. Available:
  \url{http://arxiv.org/abs/1904.04631}
\BIBentrySTDinterwordspacing

\bibitem{van2016wavenet}
A.~Van Den~Oord, S.~Dieleman, H.~Zen, K.~Simonyan, O.~Vinyals, A.~Graves,
  N.~Kalchbrenner, A.~Senior, and K.~Kavukcuoglu, ``Wavenet: A generative model
  for raw audio.'' in \emph{SSW}, 2016, p. 125.

\bibitem{huang2008three}
C.~Huang and M.~Akagi, ``A three-layered model for expressive speech
  perception,'' \emph{Speech Communication}, vol.~50, no.~10, pp. 810--828,
  2008.

\bibitem{Li2016MultilingualSE}
X.~Li and M.~Akagi, ``Multilingual speech emotion recognition system based on a
  three-layer model,'' in \emph{INTERSPEECH}, 2016.

\bibitem{Zhu_2017_ICCV}
J.~Zhu, T.~Park, P.~Isola, and A.~Efros, ``Unpaired image-to-image translation
  using cycle-consistent adversarial networks,'' in \emph{IEEE ICCV}, Oct 2017.

\bibitem{morise2016world}
M.~Morise, F.~Yokomori, and K.~Ozawa, ``World: a vocoder-based high-quality
  speech synthesis system for real-time applications,'' \emph{IEICE Trans. on
  Information and Systems}, vol.~99, no.~7, pp. 1877--1884, 2016.

\bibitem{DBLP:conf/slt/KameokaKTH18}
H.~Kameoka, T.~Kaneko, K.~Tanaka, and N.~Hojo, ``Stargan-vc: non-parallel
  many-to-many voice conversion using star generative adversarial networks,''
  in \emph{2018 {IEEE} Spoken Language Technology Workshop, {SLT} 2018, Athens,
  Greece, December 18-21, 2018}, 2018, pp. 266--273.

\bibitem{dauphin2017language}
Y.~Dauphin, A.~Fan, M.~Auli, and D.~Grangier, ``Language modeling with gated
  convolutional networks,'' in \emph{ICML}, 2017, pp. 933--941.

\bibitem{Gao18}
\BIBentryALTinterwordspacing
J.~Gao, Y.~Xu, J.~Barreiro-Gomez, M.~Ndong, M.~Smyrnakis, and H.~Tembine,
  ``Distributionally robust optimization,'' in \emph{Optimization Algorithms},
  J.~Valdman, Ed.\hskip 1em plus 0.5em minus 0.4em\relax Rijeka: IntechOpen,
  2018, ch.~1. [Online]. Available:
  \url{https://doi.org/10.5772/intechopen.76686}
\BIBentrySTDinterwordspacing

\bibitem{BurkhardtPRSW05}
F.~Burkhardt, A.~Paeschke, M.~Rolfes, W.~F. Sendlmeier, and B.~Weiss, ``A
  database of german emotional speech,'' in \emph{{INTERSPEECH} 2005 -
  Eurospeech, 9th European Conference on Speech Communication and Technology,
  Lisbon, Portugal, September 4-8, 2005}, 2005, pp. 1517--1520.

\bibitem{10.1371/journal.pone.0196391}
S.~R. Livingstone and F.~A. Russo, ``The ryerson audio-visual database of
  emotional speech and song (ravdess): A dynamic, multimodal set of facial and
  vocal expressions in north american english,'' \emph{PLOS ONE}, vol.~13,
  no.~5, pp. 1--35, 05 2018.

\bibitem{DBLP:journals/corr/UlyanovVL16}
D.~Ulyanov, A.~Vedaldi, and V.~S. Lempitsky, ``Instance normalization: The
  missing ingredient for fast stylization,'' \emph{CoRR}, vol. abs/1607.08022,
  2016.

\bibitem{Huang_2017_ICCV}
X.~Huang and S.~Belongie, ``Arbitrary style transfer in real-time with adaptive
  instance normalization,'' in \emph{ICCV}, Oct 2017.

\bibitem{DBLP:journals/taslp/TaoKL06}
J.~Tao, Y.~Kang, and A.~Li, ``Prosody conversion from neutral speech to
  emotional speech,'' \emph{{IEEE} Trans. Audio, Speech {\&} Language
  Processing}, vol.~14, no.~4, pp. 1145--1154, 2006.

\end{thebibliography}

% \begin{thebibliography}{9}
% \bibitem[1]{Davis80-COP}
%   S.\ B.\ Davis and P.\ Mermelstein,
%   ``Comparison of parametric representation for monosyllabic word recognition in continuously spoken sentences,''
%   \textit{IEEE Transactions on Acoustics, Speech and Signal Processing}, vol.~28, no.~4, pp.~357--366, 1980.
% \bibitem[2]{Rabiner89-ATO}
%   L.\ R.\ Rabiner,
%   ``A tutorial on hidden Markov models and selected applications in speech recognition,''
%   \textit{Proceedings of the IEEE}, vol.~77, no.~2, pp.~257-286, 1989.
% \bibitem[3]{Hastie09-TEO}
%   T.\ Hastie, R.\ Tibshirani, and J.\ Friedman,
%   \textit{The Elements of Statistical Learning -- Data Mining, Inference, and Prediction}.
%   New York: Springer, 2009.
% \bibitem[4]{YourName17-XXX}
%   F.\ Lastname1, F.\ Lastname2, and F.\ Lastname3,
%   ``Title of your INTERSPEECH 2019 publication,''
%   in \textit{Interspeech 2019 -- 20\textsuperscript{th} Annual Conference of the International Speech Communication Association, September 15-19, Graz, Austria, Proceedings, Proceedings}, 2019, pp.~100--104.
% \end{thebibliography}

\end{document}